\documentclass[conference, 10pt]{IEEEtran}

\usepackage{cite}
\usepackage{url}
\usepackage{graphicx}
\usepackage{color}
\usepackage{placeins}
\usepackage{float}
\usepackage{gensymb}
\usepackage{booktabs}
\usepackage{float}
\usepackage{caption}
\usepackage{subcaption}

\usepackage{multicol}
\usepackage{subcaption}
\usepackage{tabularx,colortbl}
\usepackage{ifthen}
\usepackage{graphicx}
\usepackage{caption}
\usepackage{subcaption}

\makeatletter

\newcounter{author}
\renewcommand{\author}[2][]{
   \stepcounter{author}
   \@namedef{author@\theauthor}{#2}
   \@namedef{authorlabel@\theauthor}{#1}
}

\newcounter{address}
\newcommand{\address}[2][]{
   \stepcounter{address}
   \@namedef{address@\theaddress}{#2}
   \@namedef{addresslabel@\theaddress}{#1}
}

\newcommand{\alsep}{and}

\def\newmaketitle{\par%
  \begingroup%
  \normalfont%
  \def\thefootnote{}
  \def\footnotemark{}
  \let\@makefnmark\relax
  \footnotesize
  \footnotesep 0.7\baselineskip
  \normalsize%
  \twocolumn[\thenewmaketitle\@IEEEaftertitletext]%
  \if@IEEEusingpubid
     \enlargethispage{-\@IEEEpubidpullup}%
  \fi
  \endgroup
  \setcounter{footnote}{0}\let\maketitle\relax\let\@maketitle\relax
  \gdef\@thanks{}%
  \let\thanks\relax}

\def\thenewmaketitle{
  \newpage
  \begin{center}%
    \vskip0.2em{\Huge\@IEEEcompsoconly{\sffamily}\@IEEEcompsocconfonly{\normalfont\normalsize\vskip 2\@IEEEnormalsizeunitybaselineskip
   \bfseries\large}\@title\par}\vskip1.0em\par%
    \vspace{1ex}
    \newcounter{c@author}
    \newcounter{c@tmp}
    \ifthenelse{\value{author}=2}{%
      \newcommand{\liand}{ and }}{%
      \newcommand{\liand}{, and }}
    \ifthenelse{\value{address}<2}{%
      \@nameuse{author@1}%
      \stepcounter{c@author}%
      \whiledo{\value{c@author}<\value{author}}{%
        \setcounter{c@tmp}{\value{author}}%
        \addtocounter{c@tmp}{-\value{c@author}}%
        \ifthenelse{\value{c@tmp}=1}{%
          \renewcommand{\alsep}{\liand}}{\renewcommand{\alsep}{, }}%
        \stepcounter{c@author}\alsep \@nameuse{author@\thec@author}}\\%
    }
    {
      \@nameuse{author@1}${}^{(\ref{\@nameuse{authorlabel@1}})}$%
      \stepcounter{c@author}%
      \whiledo{\value{c@author}<\value{author}}{%
      \setcounter{c@tmp}{\value{author}}%
      \addtocounter{c@tmp}{-\value{c@author}}%
      \ifthenelse{\value{c@tmp}=1}{%
        \renewcommand{\alsep}{\liand}}{\renewcommand{\alsep}{, }}%
      \stepcounter{c@author}\alsep \@nameuse{author@\thec@author}%
        ${}^{(\ref{\@nameuse{authorlabel@\thec@author}})}$%
      }
    }
    \vspace{0.2ex}

    \ifthenelse{\value{address}>0}{%
      \ifthenelse{\value{address}=1}{
        {\@nameuse{address@1}}
      }
      {
        \newcounter{c@address}

        \begin{center}
        \whiledo{\value{c@address}<\value{address}}
        {
          \refstepcounter{c@address}
            ${}^{(\thec@address)}$\,%
              \label{\@nameuse{addresslabel@\thec@address}}%
              \@nameuse{address@\thec@address}\\ %
        }
        \end{center}
      } 
    }
    {
      \relax
    }
  \end{center}
}

\makeatother

\title{Contactless Human Activity Recognition using Deep Learning with Flexible and Scalable Software Define Radio}

\author[org1]{Muhammad Zakir Khan}
\author[org2]{Jawad Ahmad}
\author[org3]{Wadii Boulila}
\author[org2]{Matthew Broadbent}
\author[org4]{Syed Aziz Shah}
\author[org3]{Anis Koubaa}
\author[org1]{Qammer H. Abbasi}

\address[org1]{James Watt School of Engineering, University of Glasgow, Glasgow, UK\\
Emails: m.khan.6@research.gla.ac.uk, Qammer.Abbasi@glasgow.ac.uk}
\address[org2]{School of Computing, Edinburgh Napier University, UK \\
Email: J.Ahmad@napier.ac.uk, m.broadbent@napier.ac.uk}
\address[org3]{Robotics and Internet-of-Things Laboratory, Prince Sultan University, Riyadh, Saudi Arabia\\
Emails: wboulila@psu.edu.sa, akoubaa@psu.edu.sa}
\address[org4]{Research Centre for Intelligent Healthcare, Coventry University, UK \\
Email: syed.shah@coventry.ac.uk}

\begin{document}

\newmaketitle
\begin{abstract}

Ambient computing is gaining popularity as a major technological advancement for the future. The modern era has witnessed a surge in the advancement in healthcare systems, with viable radio frequency solutions proposed for remote and unobtrusive human activity recognition (HAR). Specifically, this study investigates the use of Wi-Fi channel state information (CSI) as a novel method of ambient sensing that can be employed as a contactless means of recognizing human activity in indoor environments. These methods avoid additional costly hardware required for vision-based systems, which are privacy-intrusive, by (re)using Wi-Fi CSI for various safety and security applications. During an experiment utilizing universal software-defined radio (USRP) to collect CSI samples, it was observed that a subject engaged in six distinct activities, which included no activity, standing, sitting, and leaning forward, across different areas of the room. Additionally, more CSI samples were collected when the subject walked in two different directions. This study presents a Wi-Fi CSI-based HAR system that assesses and contrasts deep learning approaches, namely convolutional neural network (CNN), long short-term memory (LSTM), and hybrid (LSTM+CNN), employed for accurate activity recognition. The experimental results indicate that LSTM surpasses current models and achieves an average accuracy of 95.3\% in multi-activity classification when compared to CNN and hybrid techniques. In the future, research needs to study the significance of resilience in diverse and dynamic environments to identify the activity of multiple users.

\end{abstract}

\section{Introduction}
\par
\let\thefootnote\relax\footnotetext{This work is supported by Prince
Sultan University in Saudi Arabia} 

In recent years, there has been a notable surge in the interest surrounding HAR, mainly due to the increasing need for monitoring human behavior and activities within indoor environments. This has resulted in the development of numerous applications, including activity-assisted living for individuals with health conditions or impairments, sports, augmented reality treatment, as well as a focus on fall prevention for the elderly and disabled. HAR systems may be classified into three major categories depending on the type of sensor being used to collect data on human behaviour: radio frequency (RF)-based systems \cite{wang2017device}, wearable sensor-based systems \cite{tahir2020novel}, and vision-based systems \cite{cohen2003inference}. The vision-based systems use cameras to capture user activity, which is then analyzed using image/video processing algorithms. Despite improvements in camera sensing and image processing techniques, these systems still encounter various challenges such as the requirement for unimpeded view of the environment, lower performance in low light, and privacy concerns due to recorded video. In contrast, wearable sensor-based systems rely on inertial measurement units (IMUs) to measure and collect data on human activities. These systems are more affordable, private, and light-resistant than vision-based systems. However, they do have limitations, including the need for users to wear sensors, which may be uncomfortable for elderly/children or those with disabilities. To ensure accurate results, wearable sensor-based devices must be mounted on the body in accordance with the manufacturer's recommendation.

RF-based approaches have recently been intensively used by researchers to capture the fluctuations carried on by human activity. RF-based systems operate on the principle that RF signals reflect off human bodies, and alter in ambient RF signals occur due to human activity \cite{wang2017device, boulila2021noninvasive}. These RF-based systems require a transmitter and receiver to be installed and configured to continuously send and receive RF signals. RF-based systems are distinct from vision and wearable sensor-based systems because they do not require users to wear sensors or be concerned about their privacy. They are appropriate for recognizing indoor activities in healthcare system. These systems either utilize radar \cite{adib2014multi}, Infrared sensing \cite{rehman2021infrared} or Wi-Fi-based \cite{wei2019real} sensing techniques. Radar-based HAR systems possess a high bandwidth, resulting in a more precise spatial resolution, making them ideal for detecting fine-grained human activities. However, the expensive hardware configuration required by radar-based systems limits their widespread application. In contrast, HAR systems that rely on Wi-Fi do not require specialized hardware and can be easily incorporated into existing Wi-Fi infrastructures in various indoor environments such as homes, offices, and public places.

\par
The ubiquity of Wi-Fi routers in modern homes has made it feasible to utilize RSSI, which is the received signal strength, and CSI to sense the environment, presenting a challenging yet effective opportunity. This study showcases the potential of utilizing Wi-Fi CSI with DL for HAR. A Wi-Fi CSI and deep learning (DL) based system can improve security measures by detecting individuals in complete darkness, detecting falls in elderly people, identifying suspicious activity, and triggering rescue efforts. However, RSSI is very unstable, varies from subject to subject \cite{adib20143d}, and cannot identify signal fluctuations brought on by human activities, particularly if a person is not standing precisely between an access point and a Wi-Fi router. Regarding the channel state, the CSI function provides more specific data. It offers measurements of the amplitude and phase distortions for various sub-channels for each antenna pair of the transmitter and receiver at each sub-carrier frequency.
According to the study \cite{yousefi2017survey}, CSI can provide detailed information and is monitored for each orthogonal frequency division multiplexing (OFDM) packet, whereas RSSI offers general information. CSI can be utilized to identify human activity by analyzing the amplitude fluctuations of RF signals in the context of Wi-Fi \cite{ma2019wifi}.
\par
The aim of this research is to capture CSI information on a single subject performing activities in different regions within a room. This will be achieved by utilizing two USRP ($X300/X310$), with $X300$ acting as a transmitter ($T_x$) and the $X310$ as a receiver ($R_x$). The changes in CSI amplitude have the ability to differentiate between different activities, which can be leveraged to recognize human activity by detecting variations in RF signals. To classify six distinct activities in a single room, DL algorithms such as LSTM, CNN, and LSTM-CNN are applied. Furthermore, an additional class was included to detect an empty room. The contributions of this study can be summed up as using DL algorithms to accurately recognize six distinct activities in an indoor environment by making predictions based on the CSI collected from USRP devices.
The organization of the paper is as follows: the related work is described in section \ref{related}, while section \ref{data} describes the data and methods used. In section \ref{result}, the results and discussion are presented, and section \ref{conclusion} concludes the paper.

\section{Related Work} \label{related} 
The contactless healthcare has witnessed a significant surge in interest and progress towards HAR. This necessitates the development of systems that can monitor human activity and provide data, which may be important in resolving issues with physical and mental health as well as in the early detection and treatment of diseases. One of the most significant research results from Hoang et al. \cite{hoang2019recurrent} work on WiFi-based sensing for the user location, and tracking. Their study focused on indoor activity and using WiFi RSSI, which considers correlation among the RSSI measurement in a trajectory.
RSSI is a relatively simple measurement that neither the access point nor the mobile end of the system has to have any specific hardware modified. While RSSI is simple to use for HAR, however, it is affected by multiple factors such as multipath fading, substantial distortions, and instability, especially in complex environments \cite{cheng2017device}. According to Ma et al. \cite{ma2019wifi}, RSSI is a general information source that does not make use of the subcarriers of an OFDM channel. In contrast, Zhang et al. \cite{zhang2019wigrus} proposed a gesture recognition system named \textit{WiGrus} that operates on WiFi signals and uses detailed CSI to recognize different hand gestures. For this method, the Wi-Fi USRP-N210 software-defined radio technology is required, which also calls for upgraded Wi-Fi hardware.
Conversely, intrusive approaches \cite{hou2017tagbreathe} are widely employed and provide high accuracy, yet they are perceived as cumbersome and bothersome by the elderly or young children. Therefore, there is a growing demand for contactless, long-term health monitoring approaches that can efficiently detect activity in crowded environments \cite{hao2020wi}. In such cases involving multiple targets and the use of broad bandwidths and large antenna systems, radar-based solutions demonstrated precise detection and real-time health monitoring \cite{zhu2018indoor}. However, such approaches are expensive, energy-guzzling, and must be made more cost-effective and energy-efficient for quick and widespread adoption.

\par
Several artificial intelligence methods, including random forest (RFr), support vector machine (SVM), hidden Markov model (HMM), CNN, RNN, and LSTM, can be used for multi-class classification of HAR based on extracted characteristics. In a study by Yousefi \cite{yousefi2017survey}, RFr, HMM, and LSTM were applied on an open dataset that included six different activities: lying down, standing up from a chair, sitting, walking, and running. The dataset was collected using a NIC 5300 and 3 antennas, and the input feature vector consisted of a 90-dimensional CSI amplitude vector, which comprised of 3 antennas and 30 subcarriers. In order to extract features and denoise the CSI amplitude, the process entails utilizing the short-time Fourier transform (STFT) for feature extraction and principal component analysis (PCA). However, the classification results using RFr with 100 trees were unsatisfactory for recognizing sleeping, sitting, and standing activities. The features were retrieved using the STFT and DWT approaches while employing HMM. Compared to RFr, accuracy is better, but training time is longer. Although HMM has shown positive walking and running results, it cannot differentiate between lying in bed, sitting down, and standing up. Similarly, LSTM automatically and directly extracts the features from the raw CSI without pre-processing. To put it another way, the LSTM has a more extended training period than previous approaches but does not need PCA or STFT \cite{yousefi2017survey}. LSTM's reported accuracy for all activities is $>$75\% in \cite{zhang2020data}. Furthermore, the study conducted by Taylor et al. \cite{taylor2020intelligent}, utilized two USRPs ($X300$ and $X310$) by applying RFr algorithm to distinguish between sitting and standing activities. The accuracy rate of nearly $90\%$ was achieved when the proposed dataset was compared to a benchmark dataset. Another study conducted by Iqbal et al. \cite{iqbal2018indoor} applied DL-based system to distinguish between different user movement states such as forward, backward, and no movement. The model claimed accuracy of $89\%$ at a distance of $1.5$ meters, while at a distance of $2$ meters, the accuracy dropped to $74\%$. As a result, the model accuracy decreased when the subject moved away from the passive sensing system.

\section{Data and Methods} \label{data}
This section presents the experimental setup used to collect CSI data from contactless sensing devices for HAR. Additionally, it includes information on the techniques employed for data collection and pre-processing.
\subsection{Data Collection}
\begin{figure}
\centerline{\includegraphics [width=3in]{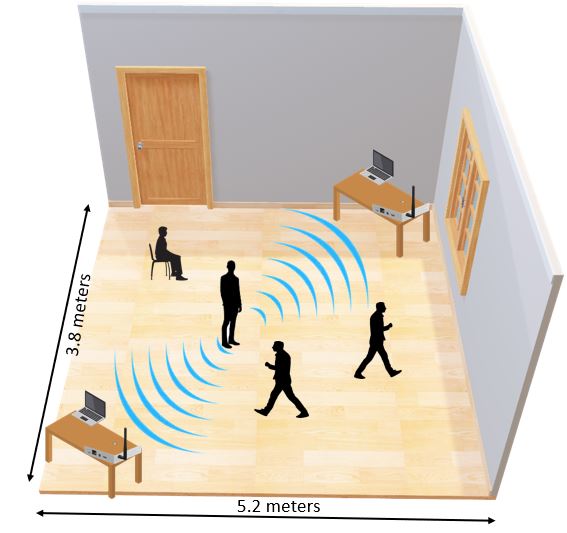}}
\caption{Experiment Setup Diagram.}
\label{ExperimentalSetupBlockDiagram}
\end{figure}

\begin{figure*}[!t]
    \begin{subfigure}[b]{0.4\columnwidth}
    \includegraphics[width=\linewidth]{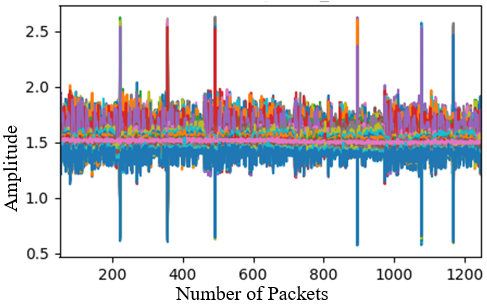}
    \caption{NoActivity}
    \label{fig:NoActivity}
  \end{subfigure}
  \begin{subfigure}[b]{0.4\columnwidth}
    \includegraphics[width=\linewidth]{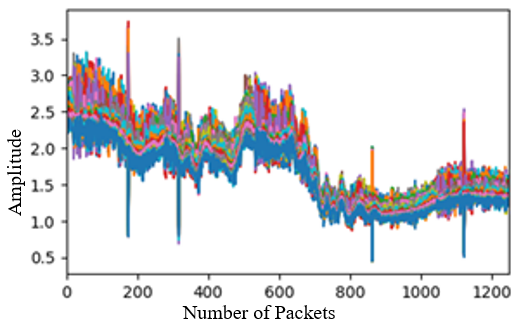}
    \caption{Sitting}
    \label{fig:Sitting}
  \end{subfigure}
    \begin{subfigure}[b]{0.4\columnwidth}
    \includegraphics[width=\linewidth]{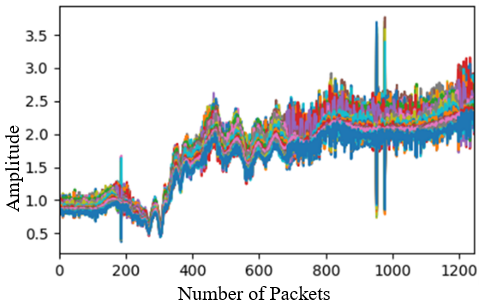}
    \caption{Standing}
    \label{fig:Standing}
  \end{subfigure}
\begin{subfigure}[b]{0.4\columnwidth}
    \includegraphics[width=\linewidth]{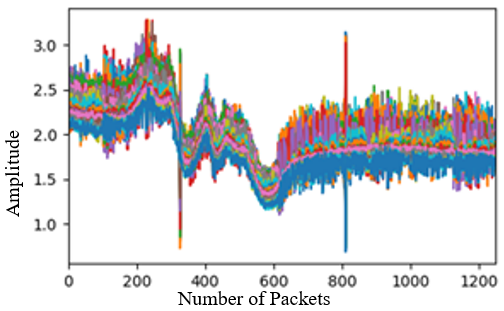}
    \caption{Leaning}
    \label{fig:Leaning}
  \end{subfigure}
\begin{subfigure}[b]{0.4\columnwidth}
    \includegraphics[width=\linewidth]{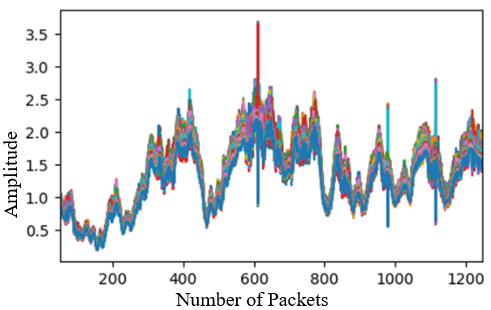}
    \caption{Walking}
    \label{fig:WalkingRxTx}
  \end{subfigure}
\caption{CSI sample of activity representing various activity classes}%
    \label{Activities}%
\end{figure*}

The Wi-Fi signal is transmitted and the CSI received utilizing two USRP devices, both of which are placed in opposing corners of a room at a $45\degree$ angle, as depicted in Fig. \ref{ExperimentalSetupBlockDiagram}. The data is collected by a single subject performing different activities in the room. The $T_x$ device utilizes $3.75 GHz$ and $64$ OFDM sub-carriers with gain levels set at $70dB$ for $T_x$ and $50dB$ for $R_x$. The GNU radio flow diagram is converted into a Python script that initiates the OFDM communication and outputs the collected CSI during transmission. Each data packet contains a total of $T_x$, $R_x$, and CSI streams. Fig. \ref{Activities} displays the six activities and empty class along with CSI amplitude fluctuations. The color of each sub-carrier represents a sub-carrier an activity, and the number of packets on the x-axis and the amplitude of the sub-carrier on the y-axis. Each data sample represents an OFDM transmission of three seconds, resulting in a sample of $1200$ packets. A total of $700$ data samples are collected, with $100$ samples for each of the seven activities.

\subsection{Pre-Processing}
\begin{figure}[H]
\centering
\includegraphics[width=3.5in]{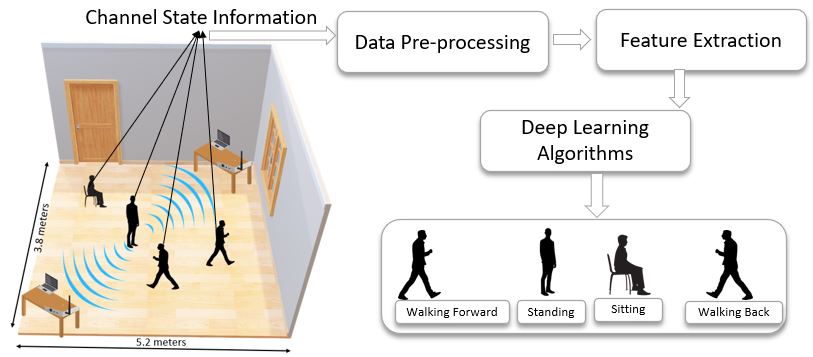}
\caption{Data Flow Diagram depicting human activity captured as CSI \& compiled into a dataset for classification.}
\label{FullSystem}
\end{figure}

The initial step in data preprocessing involves data organization, which is a key aspect of data analysis as it enables algorithms to use data as input. The handling of missing "NaN" values in datasets is important as ignoring these values can result in the loss of important information. Hence, we substituted the "NaN" values with the mean value. To prepare the raw data for analysis, we utilized the popular python data analysis library, \textit{scikit}, that is used to parse matrix (CSV) files using the \textit{pandas} package. The generated data frames from CSV files were examined using \textit{Scikit-learn}. We applied the PCA technique, specifically the \textit{fit-transform} method, to reduce dimensions. We also applied two feature selection methods, namely \textit{selectKBest} and \textit{ANOVA f-test}. A digital and analog Butterworth filter produced a \textit{Butterworth} filter signal of $n$-th order, returning the $(B, A)$ form of the filter coefficients using the \textit{butter(1, 0.05)} function. Furthermore, we added labels to the first column of the data frames. Since the label column contains categorical values, we applied label encoding to maintain the same number of features. Label-encoding results in data that is simpler to incorporate in DL algorithms and has faster processing time than one-hot encoding. The output of the python scripts was the CSI collected during transmission, which was represented by complex numbers. The absolute value of the complex numbers was used to calculate the amplitude of the signals. Afterwards, we transformed the CSI amplitude data into CSV files, which were utilized for training and testing algorithms. Fig. \ref{FullSystem} displays the data flow diagram for the data collection activities related to the seven activity classes.

\subsection{Human Activity using Deep Neural Networks}
Nowadays, deep learning methods are employed extensively to extract features automatically from the impact of activities on CSI. While using a neural network with many layers can improve classification accuracy, overfitting and performance deterioration can become significant when dealing with small amounts of data. Traditional techniques such as learning rate, small batch size, and weight decay may not suffice in preventing these problems. As a result, in order to achieve the desired performance, each of the existing WiFi-based systems needs to integrate a specific number of dedicated neural layers. In this study, we have utilized DL models, which are more effective when dealing with small datasets and need less processing time.

\subsubsection{Human Activity using LSTM}
As an alternative approach, we have utilized LSTM networks. These are artificial RNNs, referred to as LSTM, which are ideal for handling time-series data and applicable to our situation, as reported in \cite{santos2021artificial,al2021novel}. Our focus for the LSTM algorithm has been limited to extracting CSI values and reducing noise. Each of the $64$ subcarriers' raw CSI amplitude is incorporated into a $64$-dimensional feature vector. The hidden layer has a size of $(20,50)$, with \textit{tanh} activation being used. To minimize cross-entropy loss, Adam optimizer is used with a batch size of $64$, a learning rate of $0.01$, and decay of $1e{^{-6}}$. One benefit of using LSTM for classification is that it allows for direct learning from raw series data, which eliminates the requirement of manually engineering input features and grants more flexibility to domain experts.

\subsubsection{Human Activity using CNN}
The CNN is one of the most popular DL architectures due to its ability to automatically extract deep, high-dimensional features as compared to just a few shallow ones \cite{ben2022randomly,ben2022fusion}. We contend that CNN and LSTM are compatible DL techniques based on the work \cite{greff2016lstm}. Although LSTM is better suited for time-domain analysis and responds better to short-duration movement, CNN focuses primarily on changes in the frequency domain and has a greater reaction to long-duration movement. The size of the hidden layers was maintained at $(20,32)$, the maximum pooling size was set at $(3,1)$, and the activation is \textit{tanh}. At this stage, we can make the final classification prediction in the output layer by adopting the general CNN design principle.

\subsubsection{Human Activity using LSTM-CNN}
The LSTM is an RNN variant that works well with continuous temporal relationships and is appropriate for processing time series. CNN can reduce frequency domain changes and extract spatial features \cite{hammerla2016deep}. The benefits of the two models are combined in the LSTM-CNN model. It is important that different LSTM variations provide different results when used as the input for CSI data. We contrast LSTM with LSTM-CNN to identify human activities. The LSTM layer analyzes the input. The changes in signal fluctuation are utilized for collecting the input data, followed by the extraction of time-domain characteristics from the initial signal. The 1D-CNN layer is applied after the LSTM's output. Convolution allows for the extraction of high-dimensional implicit features. The optimal feature sequence is then obtained by passing the convolutional results through a maximum pooling layer, which is a direct foundation for the final classification. We have utilized LSTM hidden layer size $(20,50)$, CNN hidden layer size $(50,32)$, 1D max-pooling-size is $(3,1)$, and activation is \textit{tanh} and \textit{softmax} as mentioned in Table \ref{hyperparameters}.

\begin{table}
\caption{Hyper-parameters of deep learning algorithms}
\begin{center}
\begin{tabular}{|p{0.4cm}|p{1.5cm}|p{5.5cm}|}
\hline \textbf{S.No} & \textbf{Algorithms} & \textbf{Hyper Parameters}  \\
\hline 1 & LSTM & optimizer='adam', Activation  = tanh, lr=(0.1,0.01), decay= 1$e^{-6}$, epochs=(20,50), hidden-layer-size=(20,50), Dropout=0.2 , batch-size=32, connected-layer-activation:'softmax' \\
\hline 2 & CNN & optimizer='adam', Activation  = tanh, lr=(0.1,0.01), loss = 'binary-crossentropy', hidden-layer-size=(20,32), epochs=(20,50), batch-size=32, Max-pooling-size = (3,1), connected-layer-activation:'softmax' \\
\hline 3 & LSTM-CNN &  optimizer='adam', Activation  = tanh, lr=(0.1,0.01), loss = 'binary-crossentropy', epochs=(20,50), LSTM hidden-layer-size= (20,50), CNN hidden-layer-size= (50,32),
Max-pooling-size = (3,1)
  \\
\hline
\end{tabular}
\label{hyperparameters}
\end{center}
\end{table}

\section{Result} \label{result}
\subsection{Deep Learning Parameters and evaluation metrics}
In this section, we evaluate the model's performance for multi-classifications using several metrics, including accuracy, loss, and confusion matrices. In the experiment, 700 samples of data were collected for training and testing, comprising two groups of static data (Empty and No activity) and 500 samples of dynamic data (sitting, standing, leaning forward, and walking in two directions).

The three important parameters in DL are the number of \textit{epochs}, which dictates how many times the model processes the entire training dataset; the \textit{batch size}, which determines how much training data to process before updating the model's internal parameters, and the \textit{learning rate}, which governs the amount of change in the model weights following an estimated error update. Our experimental results indicate that modifying the batch size to 32 and reducing the vector learning rate to 0.01 gives a noteworthy enhancement in the network's classification performance.

Our results show that vector selection is important to network performance and that hyper-parameter optimization can increase classification accuracy. Table \ref{classification} shows that after 50 iterations of varying experimental conditions, the rate of change shows minimal variation while keeping a training size of 0.80. The Keras DL library supports the \textit{scikit-learn} module's \textit{make\_multilabel\_classification()} function and utilizes the \textit{ReLU} and \textit{softmax} activation functions in the hidden layer.

\begin{table}[H]
\caption{Classification accuracy of deep learning algorithms}
\begin{center}
\begin{tabular}{|p{0.4cm}|p{1.5cm}|p{1cm}| p{1.1cm}|p{1cm}|p{1cm}|}
\hline \textbf{S.No} & \textbf{Algorithms} & \textbf{Accuracy} & \textbf{F1 Score} & \textbf{Precision} & \textbf{Recall} \\
\hline 1 & LSTM & 95.3\% & 0.78 & 0.82 & 0.74 \\
\hline 2 & LSTM-CNN & 91.0\% & 0.60 & 0.85 & 0.46 \\
\hline 2 & CNN & 86.5\% & 0.20 & 0.40 & 0.14 \\
\hline
\end{tabular}
\label{classification}
\end{center}
\end{table}

\subsection{Comparison with other models}
\begin{figure}
\centerline{\includegraphics [width=3.2in]{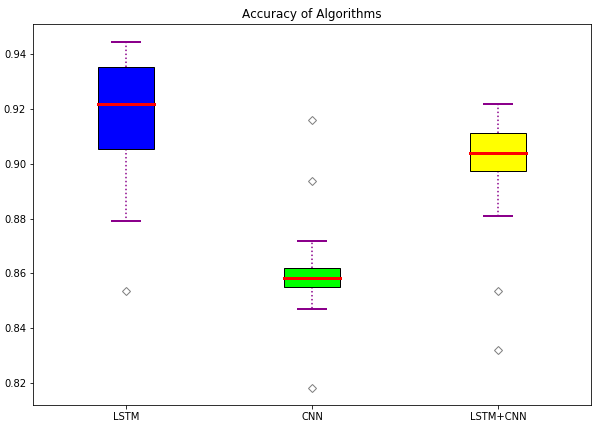}}
\caption{Comparison of deep learning Algorithms.}
\label{Comparison}
\end{figure}

To evaluate the classification performance, we compared the LSTM network with that of the LSTM-CNN and CNN networks. Both of them use the TensorFlow library in the Python environment, and the number of LSTM hidden layers they employ matches the input data. The LSTM network has a better classification result, as demonstrated in Fig. \ref{Comparison}. The average loss value for multi-classification is 12.5\%, the average accuracy rate is 95\%, and all indices outperform LSTM-CNN which has 91\%, and CNN which has 86\%. The average loss values for LSTM-CNN and CNN are 20.3\% and 27.3\% respectively. The learning rate and epochs are slightly changed, leading to accuracy degradation, as shown in Table \ref{parameters}. In conclusion, compared to other algorithms in the same family, the suggested LSTM network provides a straightforward, effective, and highly accurate approach for recognizing seven different activities.

\begin{table}[!htbp]
\centering
\caption{Learning rate vs. epochs.}
\begin{tabular}{|c|c|c|c|c|}
\toprule
\hline
\textbf{epochs= 50} &  \multicolumn{2}{c|}{\textbf{lr = 0.01}  } & \multicolumn{2}{c|}{\textbf{lr = 0.1}} \\
\hline
\midrule
 Algorithms  & Accuracy   & Loss    & Accuracy   & Loss\\
\hline
\textbf{LSTM}   &  95.3\% & 12.5   & 93.4\%  & 14.7\\

\textbf{LSTM-CNN}   &  91.0\% & 20.3   & 75\%  & 1.85\\
\textbf{CNN}   &  86.5\%  &  27.3   & 77\%  & 1.59\\
\hline
\textbf{epochs= 20} &  \multicolumn{2}{c|}{\textbf{lr = 0.01}  } & \multicolumn{2}{c|}{\textbf{lr = 0.1}} \\
\hline
\midrule
Algorithms   & Accuracy   & Loss    & Accuracy   & Loss\\
\hline
\textbf{LSTM}   &  93.5\% & 11   & 89.4\%  & 23\\

\textbf{LSTM-CNN}   &  88\% & 23   & 74\%  & 1.38\\
\textbf{CNN}   &  84\%  &  31   & 74\%  & 1.61\\
\hline
\bottomrule
\end{tabular}
\label{parameters}
\end{table}

\subsection{Discussion}
\begin{figure}
     \centering
     \begin{subfigure}[b]{0.5\textwidth}
         \centering
         \includegraphics[width=\textwidth]{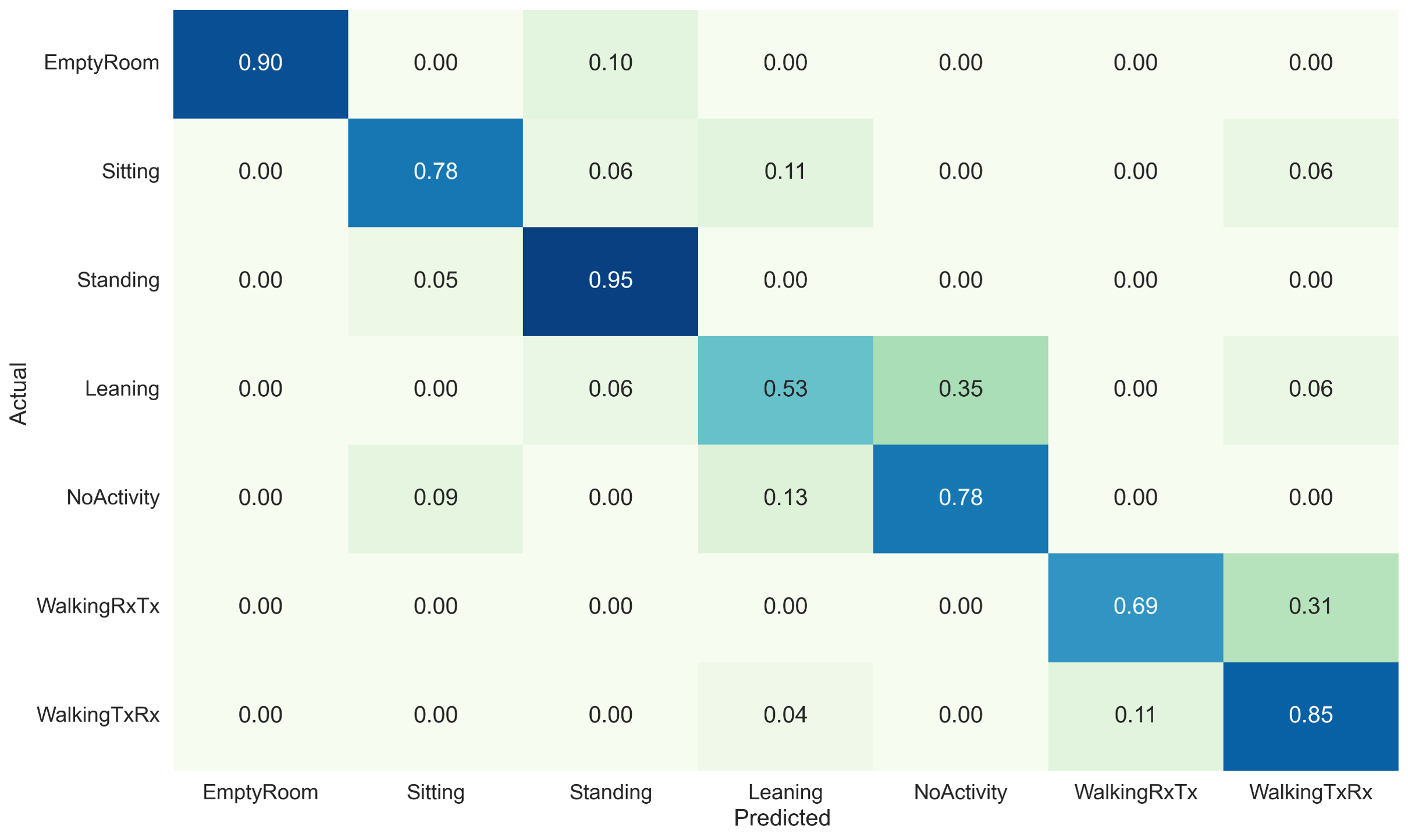}
         \caption{Normalised confusion matrix of LSTM}
         \label{fig:LSTM}
     \end{subfigure}
     \hfill
     \begin{subfigure}[b]{0.5\textwidth}
         \centering
         \includegraphics[width=\textwidth]{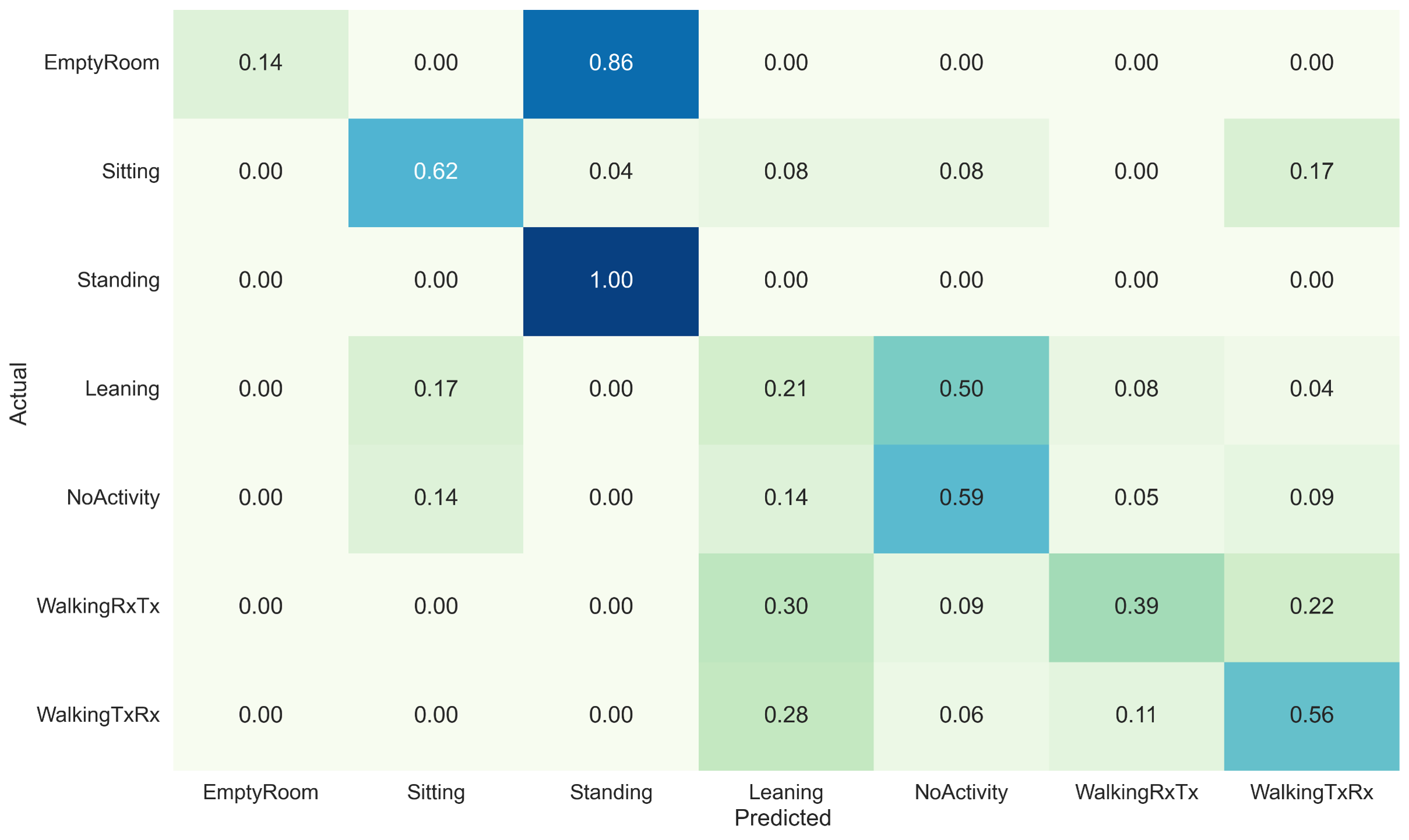}
         \caption{Normalised confusion matrix of CNN}
         \label{fig:CNN}
     \end{subfigure}
     \hfill
     \begin{subfigure}[b]{0.5\textwidth}
         \centering
         \includegraphics[width=\textwidth]{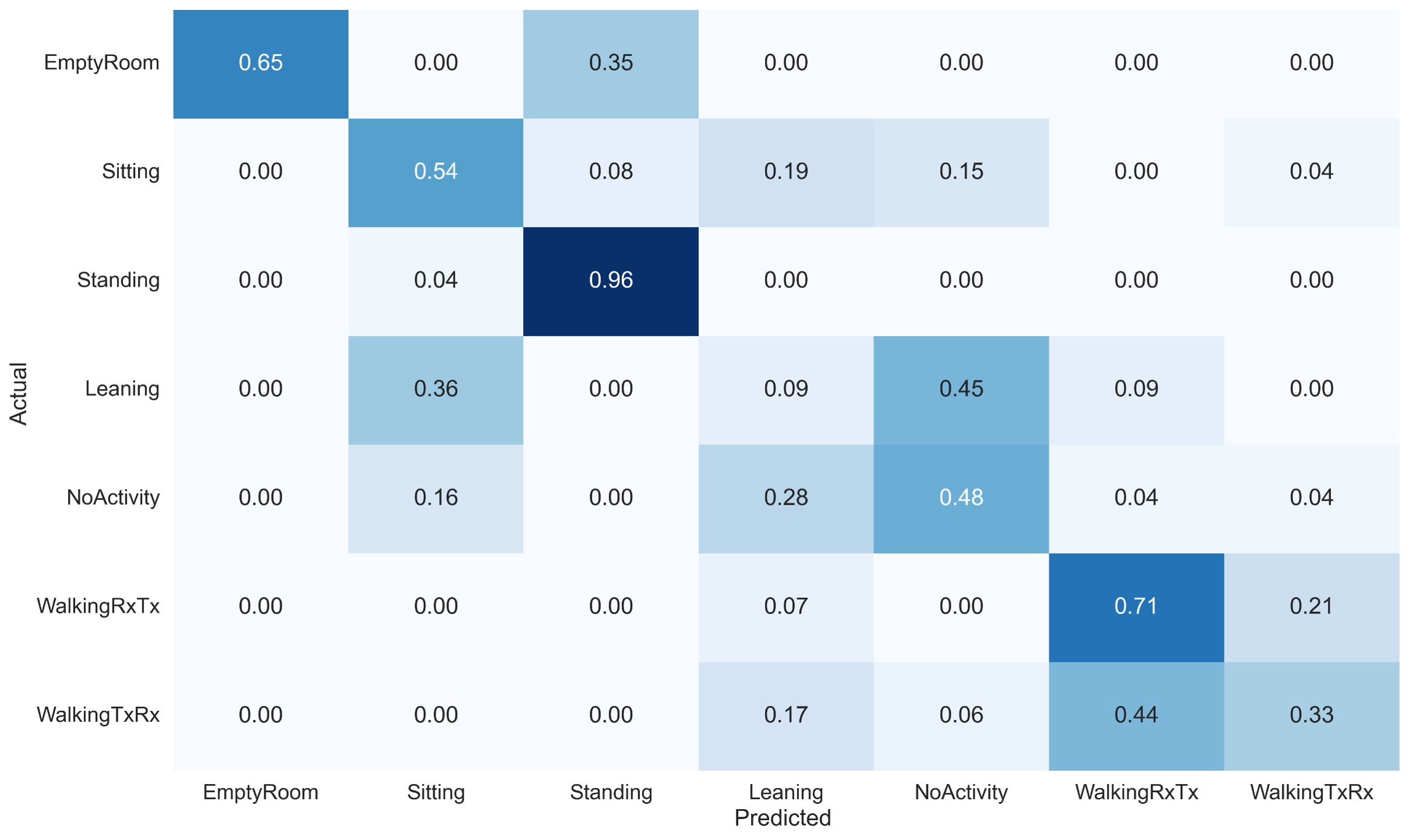}
         \caption{Normalised confusion matrix of LSTM+CNN}
         \label{fig:LSTM+CNN}
     \end{subfigure}
        \caption{The deep learning model's normalized confusion matrix for multi-classification.}
        \label{fig:confusion}
\end{figure}

The confusion matrix of the final results of the LSTM, CNN, and LSTM-CNN models on multi-classifications is shown in Figure \ref{fig:confusion}. The classification performance generated by the "empty" and "standing" activities is optimal. However, the classification performance of "walking" is poor. It is evident that certain "walking" actions are subdivided into "forward and backward," most likely because the two signals don't have clear classification criteria and have more comparable amplitude and frequency. Overall, the classification accuracy rate remains steady at above 95\%, indicating a positive classification performance.

\section{Conclusion} \label{conclusion}
The study compares the performance of the LSTM, CNN, and LSTM-CNN network models for RF-based indoor HAR. The research focuses on recognizing seven distinct activities in a single room using CSI data inputs. The results indicate that RF sensing is a viable contactless way to recognize human activities. We preprocessed the data, extracted the feature set, and implemented a complete CSI data collection system. The LSTM model outperforms the other models in terms of classification accuracy and has higher performance when accurately extracting hidden features from the CSI data. These results have a significant impact on the CSI data's HAR and highlight the viability of RF sensing for indoor activity recognition.


\bibliography{references}
\bibliographystyle{ieeetr}

\end{document}